\pdfoutput=1

\documentclass[11pt]{article}

\usepackage[]{ACL2023}

\usepackage{times}
\usepackage{latexsym}
\usepackage[normalem]{ulem}

\usepackage[T1]{fontenc}

\usepackage[utf8]{inputenc}

\usepackage{microtype}

\usepackage{inconsolata}
\usepackage{multirow}
\usepackage{graphicx}
\usepackage{adjustbox}
\usepackage{amsmath}
\usepackage{amssymb}
\usepackage[cjk]{kotex}
\usepackage{booktabs}
\usepackage{comment}
\usepackage{makecell}
\usepackage{xcolor}
\usepackage{caption}

\usepackage{floatrow}
\usepackage{graphicx}
\usepackage{subcaption}

\newcommand{\ours}{\texttt{CEAA}}
\title{Cross Encoding as Augmentation: \\Towards Effective Educational Text Classification}

\author{
\quad Hyun Seung Lee\thanks{\quad Equal Contribution.}~~$^{1, 2}$
\quad Seungtaek Choi\footnotemark[1]~~\thanks{\quad Corresponding authors.}~~$^{1}$\\
\quad \textbf{Yunsung Lee}$^{1}$ 
\quad \textbf{Hyeongdon Moon}$^{1}$
\quad \textbf{Shinhyeok Oh}$^{1}$ \\
\quad \textbf{Myeongho Jeong}$^{1}$ 
\quad \textbf{Hyojun Go}$^{1}$
\quad \textbf{Christian Wallraven}\footnotemark[2]~~$^{2}$\\
$^1$Riiid AI Research\\
$^2$Department of Artificial Intelligence, Korea University \\
\texttt{\{hyunseung.lee, seungtaek.choi\}@riiid.co}, \\
\texttt{\{hslrock, wallraven\}@korea.ac.kr}\\
}

\begin{document}
\maketitle

\begin{abstract}
\label{sec:abstract}
Text classification in education, usually called \textit{auto-tagging}, is the automated process of assigning relevant tags to educational content, such as questions and textbooks.  
However, auto-tagging suffers from a data scarcity problem, which stems from two major challenges: 1) it possesses a large tag space and 2) it is multi-label. 
Though a retrieval approach is reportedly good at low-resource scenarios, there have been fewer efforts to directly address the data scarcity problem.
To mitigate these issues, here we propose a novel retrieval approach \ours~that provides effective learning in educational text classification. Our main contributions are as follows: 1) we leverage transfer learning from question-answering datasets, and 2) we propose a simple but effective data augmentation method introducing cross-encoder style texts to a bi-encoder architecture for more efficient inference.
An extensive set of experiments shows that our proposed method is effective in multi-label scenarios and low-resource tags compared to state-of-the-art models.

\end{abstract}
\section{Introduction}
\label{sec:introduction}

Due to the overwhelming amount of educational content available, students and teachers often struggle to find what to learn and what to teach.  Auto-tagging, or text classification in education, enables efficient curation of content by automatically assigning relevant tags to educational materials, which aids in both students' understanding and teachers' planning~\cite{goel2022k}. 

However, applying auto-tagging for real-world education is challenging due to \textbf{data scarcity}. 
This is because auto-tagging has a potentially very large label space, ranging from subject topics to knowledge components (KC)~\cite{zhang2015character,koedinger2012knowledge,mohania2021tagrec,viswanathan2022tagrec++}. The resulting data scarcity decreases performance on rare labels during training~\cite{chalkidis2020empirical, lu2020multi, snell2017prototypical,choi2022c2l}.

In this paper, we aim to solve the data scarcity problem by formulating the task as a retrieval problem following a recent proposal~\cite{viswanathan2022tagrec++}. 
This can utilize a language model's ability to understand the tag text, such that 
even for an unseen tag, the models would be able to capture the relationship between the terms in the input content and labels.
However, performance in the auto-tagging context still critically depends on the amount of training data.

\begin{figure*}[t]
     \centering
     
     \begin{subfigure}[t]{0.26\textwidth}
         \centering
         \includegraphics[width=\textwidth,height=4cm]{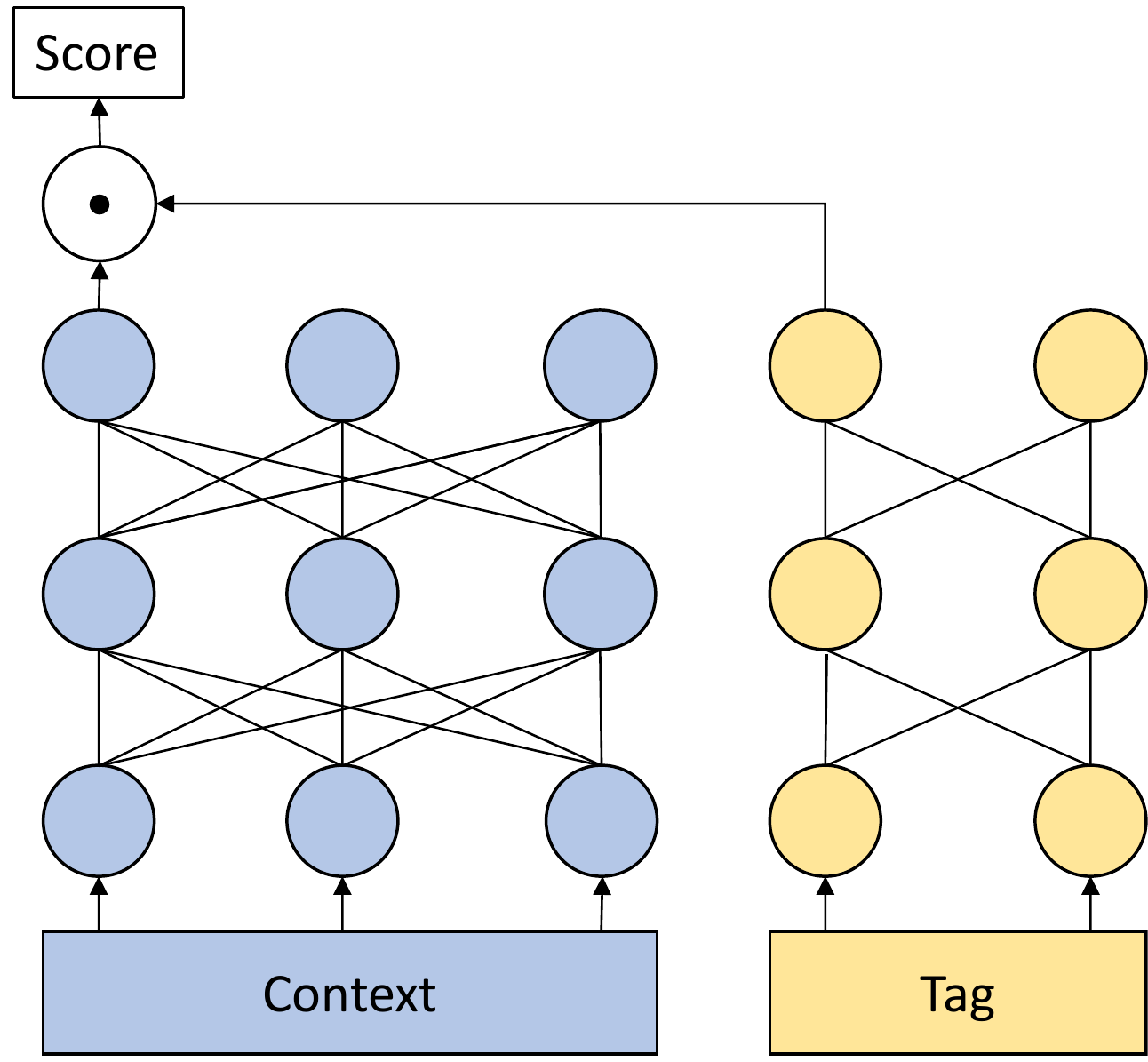}
        \label{fig:bi-encoder}
        \vspace{-\baselineskip}
        \caption{Bi-Encoder}  
        
    \end{subfigure}
     \hfill
     \begin{subfigure}[t]{0.28\textwidth}
         \centering
         \includegraphics[width=\textwidth,height=4cm]{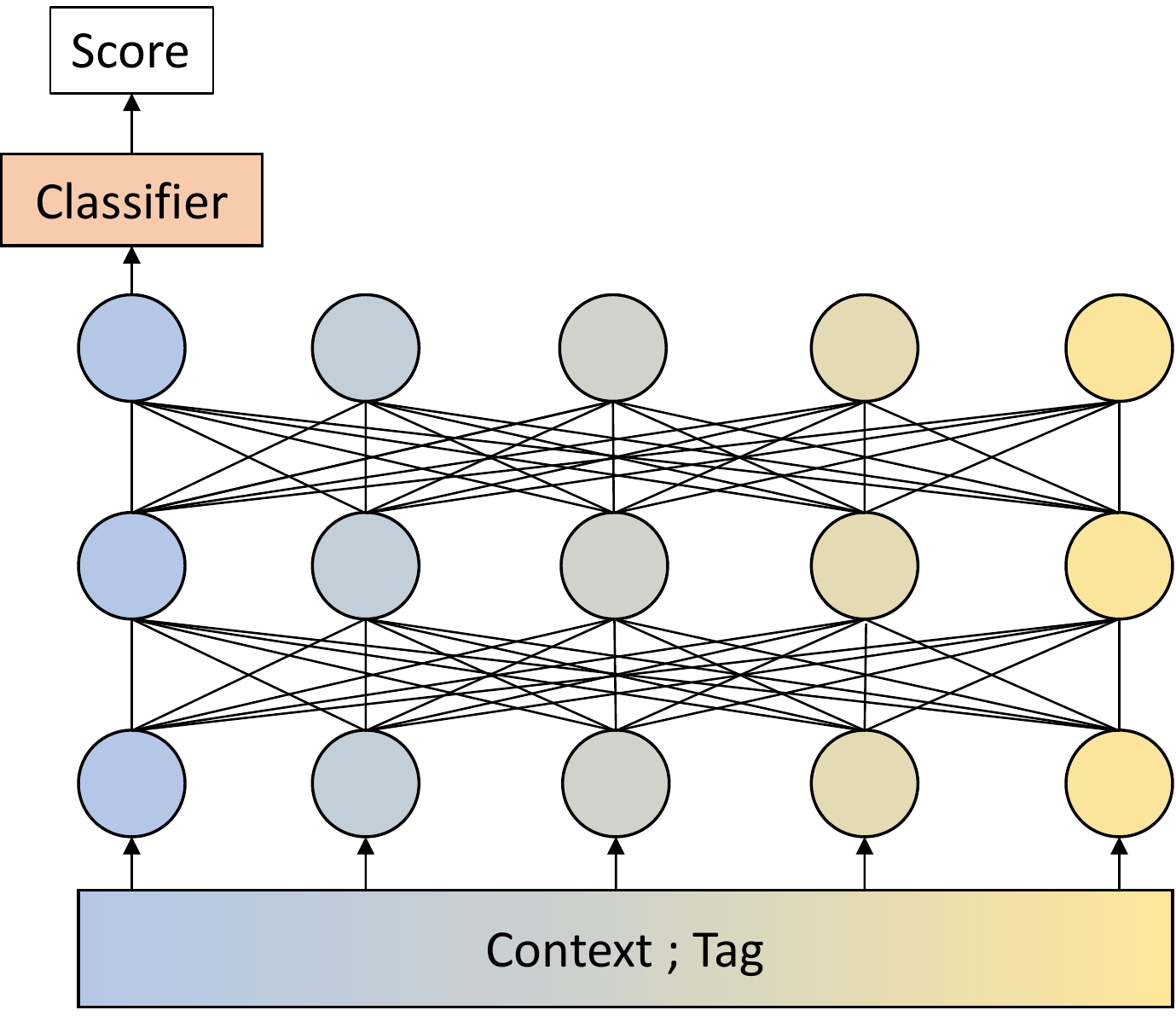}
            \caption{Cross-Encoder}
            \label{fig:cross-encder}
    \end{subfigure}
     \hfill
     \begin{subfigure}[t]{0.40\textwidth}
         \centering
         \includegraphics[width=\textwidth,height=4cm]{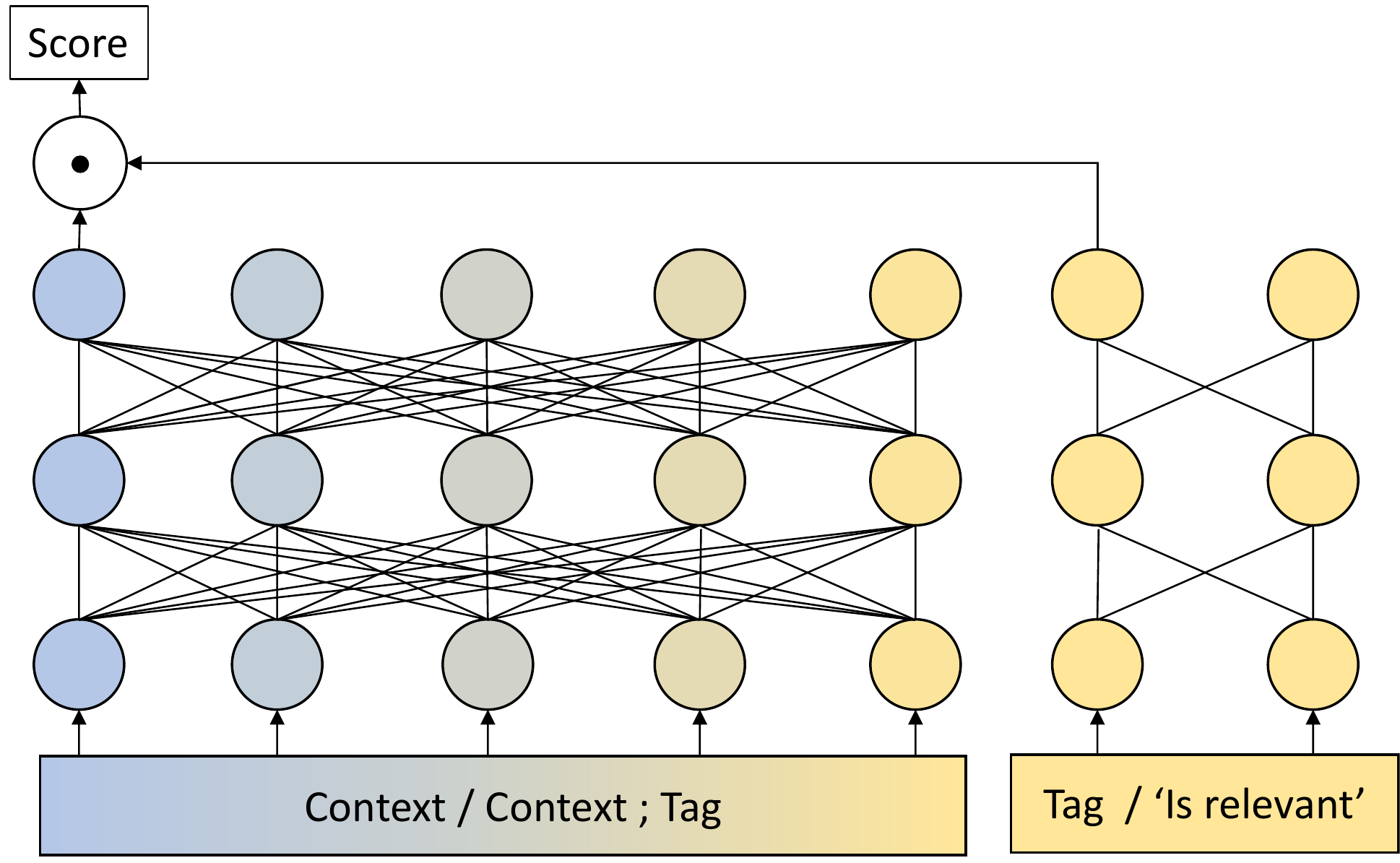}
            \caption{\ours}
            \label{fig:bi-encoder_auged}
    \end{subfigure}
    \caption{Comparative Illustration of Encoding Methods. \ours~ is done for the bi-encoder to process input in which context and tag are given together, computing full token-level interactions between context and tag. }
    \label{fig:main_diagram}
\end{figure*} 

To this end, we first propose to leverage the knowledge of language models that are fine-tuned on large question-answering datasets. 
Our intuition is that question of finding an answer in a passage can be a direct (or indirect) summary of the passage~\cite{nogueira2019document}, which can serve as an efficient proxy of the gold tag for educational content. 
The large question-answering datasets thus become a better prior for the tag spaces. 
Specifically, we adopt a recent bi-encoder architecture, called DPR~\cite{karpukhin2020dense}\footnote{DPR model is trained on 307k training questions, which is much larger than 7k questions in ARC dataset~\cite{xu2019multi} we used in experiments.}, for transfer learning, which performs BERT encoding over the input and candidate label separately and measures the similarity between the final representations. 
To the best of our knowledge, our work is the first to leverage transfer learning from QA models for text classification tasks.

As a further innovation, we introduce a novel data augmentation method for training a bi-encoder architecture, named \ours, which adds the cross-encoder \textit{view} of the input-label pair in the bi-encoder architecture, as shown in Figure 1. 
By capturing the full interaction between input and labels already during training time, the models can be further optimized to take advantage of token-level interactions that are missing in traditional bi-encoder training. At the same time, the computational efficiency of the bi-encoder is maintained, which makes \ours~able to tackle large label spaces as opposed to existing solutions based on cross-encoder architectures \cite{urbanek2019learning,wolf2019transfertransfo,vig2019comparison}. 
Experiments show that \ours~provides significant boosts to performance on most metrics for three different datasets when compared to state-of-the-art models.
We also demonstrate the efficacy of the method in multi-label settings with constraints of training only with a single label per context.
\section{Related Work}
\label{sec:related}
Text classification in the education domain is reportedly difficult as the tags (or, labels) are hierarchical~\cite{xu2019multi,goel2022k,mohania2021tagrec}, grow flexibly, and can be multi-labeled~\cite{medini2019extreme, dekel2010multiclass}. 
Though retrieval-based methods were effective for such long-tailed and multi-label datasets~\cite{zhang2022long,chang2019x}, they relied on vanilla BERT \cite{devlin2018bert} models, leaving room for improvement, for which we leverage question-answering fine-tuned retrieval models~\cite{karpukhin2020dense}. 

Recently, \cite{viswanathan2022tagrec++} proposed TagRec++ using a bi-encoder framework similar to ours, with an introduction of an additional cross-attention block. 
However, this architecture loses the efficiency of the bi-encoder architecture in the large taxonomy space for the education domain.  
Unlike TagRec++, our distinction is that we leverage the cross-attention only in training time via input augmentation. 

\section{Approach}
\label{sec:approach}

\subsection{Problem formulation}

In this paper, we address the text classification task,  which aims to associate an input text with its corresponding class label, as a retrieval problem. 
Formally, given a context $c$ and tag candidates $\mathcal{T}$, the goal of the retrieval model is to find the correct (or, relevant) tag $t \in \mathcal{T}$, where its relevance score with the context $s(c, t)$ is the highest among the $\mathcal{T}$ or higher than a threshold. 
For this purpose, our focus is to better train the scoring function $s(c, t)$ to be optimized against the given relevance score between the context $c$ and candidate tag $t$. 

\subsection{Bi-Encoder}
In this paper, we use a bi-encoder as a base architecture for the retrieval task, as it is widely used for its fast inference \cite{karpukhin2020dense}.
Specifically, the bi-encoder consists of two encoders, $E_C$, and $E_T$, which generate embedding for the context $c$ and the tag $t$. The similarity between the context and tag is measured using the dot-product of their vectors:
\begin{equation}
    s_{\text{BE}}(c, t)=E_C(c) \cdot E_T(t)^{\top}
    \label{equ:similarity_biencoder}
\end{equation}

Both encoders are based on the BERT architecture \cite{devlin2018bert}, specifically \textit{``bert-base-uncased''} provided by HuggingFace~\cite{wolf2020transformers}, that is optimized with the training objective of predicting randomly-masked tokens within a sentence.
We use the last layer's hidden layer of the classification token is used as context and tag embeddings. 

For training the bi-encoder, we follow the in-batch negative training in \cite{karpukhin2020dense}. Gold tags from other contexts inside the batch are treated as negative tags. As tags are often multi-labeled, we use \textit{binary cross-entropy loss}:
\begin{equation}
\begin{aligned}
\mathcal{L}=-\frac{1}{M}\sum_{i=1}^{M}\sum_{j=1}^{N}(y_{i,j} \log(s(c_i,t_j) \\
+(1-y_{i,j})\log(1-s(c_i,t_j))
\end{aligned}
\end{equation}
where $s(c_i,t_j)$ scores the similarity between context $c_i$ and tag $t_j$, and $y_{i,j}$ is 1 if they are relevant and 0 otherwise.
We will denote this model variant as a bi-encoder (BERT) below. 



\begin{table*}[t]
\small
    \begin{tabular}{l c c c c c c c c}
        \toprule
\multirow{2}{*}{\textbf{Methods}}
&\multicolumn{3}{c}{\textbf{ARC}}  
&\multicolumn{3}{c}{\textbf{QC-Science}}        
&\multicolumn{2}{c}{\textbf{EURLEX57K}}        

\\
 \cmidrule{2-9}
&R@1 & R@3 & R@5 & R@1 & R@3 & R@5  & RP@5 & nDCG@5
\\\hline
BM25 & 0.14 & 0.28 & 0.34 & 0.13 & 0.23 & 0.27 & 0.15 & 0.15 \\\hline

BERT (prototype) & 0.35 & 0.54 & 0.64 & 0.54 &  0.75 & 0.83 & - & 
- \\
TagRec & 0.36 & 0.55 & 0.65 & 0.54 & 0.78 & 0.86 & - & - \\ 
TagRec++ & 0.49 & 0.71 & 0.78 & 0.65 & 0.85 & 0.90 & - & - 
\\\hline
BERT (classification)  & 0.53 & 0.72 & 0.79 & 0.68 & \textbf{0.87} & \textbf{0.91} & 0.78  & 
0.80  \\

Poly-encoder-16 & 0.40 & 0.65 & 0.75 & 0.50 & 0.75 & 0.83 & 0.22 & 0.23  \\
Poly-encoder-360 & 0.44 & 0.68 & 0.78 & 0.64 & 0.85 & 0.90 & 0.54 & 0.54 \\\hline
Bi-encoder (BERT) & 0.51 & 0.71 & 0.77 & 0.67 & 0.85 & 0.90 & 0.74 & 0.76  \\
Bi-encoder (BERT) + \ours & 0.50 & 0.72 & \textbf{0.80} & 0.68 & 0.86 & 0.90 & 0.76 & 0.78 
\\

Bi-encoder (DPR)& 0.54 & 0.73 & \textbf{0.80} & 0.69 & \textbf{0.87} & 0.90 & 0.76 & 0.78  \\
Bi-encoder (DPR) + \ours & \textbf{0.56} & \textbf{0.74} & \textbf{0.80} & \textbf{0.70} & 0.86 & 0.90 & 
\textbf{0.79} & \textbf{0.81}
\\

\bottomrule
        
    \end{tabular}
    \caption{Results  of experiments on ARC, QC-Science, and EURLEX57K dataset. We mainly compared Bi-encoder with Bi-encoder + \ours~where each encoder is pretrained with different training objectives, BERT and DPR.} 
    \label{tab:main_result}    
\end{table*}








\subsection{Cross-Encoding As Augmentation}

The cross-encoder \cite{nogueira2019passage} is another method in information retrieval tasks in which a single BERT model receives two inputs joined by a special separator token as follows: 
\begin{equation}
    s_{\text{CE}}(c, t) = F(E([c; t])),
\end{equation}
where $F$ is a neural function that takes the representation of the given sequence.

Cross-encoders perform better than bi-encoders as they directly compute cross-attention over context and tag along the layers \cite{urbanek2019learning, wolf2019transfertransfo, vig2019comparison}. However, relying on this approach is impractical in our scenario as it requires processing every existing tag for a context during inference time. 
As a result, this method is typically used for \textit{re-ranking} \cite{nogueira2019multi,qu2021rocketqa,ren2021rocketqav2}. 

As shown in Figure \ref{fig:main_diagram}, we adopt an augmentation method that enables the bi-encoder framework to mimic cross-encoder's representation learning. 
Compared to other knowledge distillation methods \cite{qu2021rocketqa, ren2021rocketqav2, thakur2020augmented}, our approach does not require an additional cross-encoder network for training. 
Furthermore, as such cross-encoding is introduced as an augmentation strategy, it doesn't require additional memory or architecture modifications, while improving the test performance. 

Specifically, for a context $c$, we randomly sample one of the tags in the original batch. We extend the batch in our training by introducing a context-tag concatenated input $[c; t]$ which has ``\textit{is relevant}'' as a gold tag. Our bi-encoder must be able to classify relevance when an input includes both context and tag with the following score function:
\begin{equation}
    s_{\text{CEAA}}(c, t) = E_{C}([c; t]) \cdot E_{T}(``\text{is relevant''})^{\top}
\end{equation}

Since we use the augmentation method via input editing without an extra teacher cross-encoder model for distillation, we call this model Cross Encoding As Augmentation (\ours).


\subsection{Transfer Learning}
To overcome the data scarcity in auto-tagging tasks, we introduce bi-encoder (DPR) models that distill knowledge from large question-answering datasets. We argue that the training objective of question answering is similar to the context and tag matching in the auto-tagging task, as a question is a short text that identifies the core of a given context.
Therefore, while the previous works have relied on vanilla BERT, here we explore whether pertaining on question-answering tasks would improve the performance in the auto-tagging tasks.
Specifically, we replace the naive BERT encoders with DPR~\cite{karpukhin2020dense}, which is further optimized with the Natural Question dataset~\cite{lee2019latent,kwiatkowski2019natural} to solve open-domain question-answering tasks of matching the representations of document and question. 
To match the overall length of the texts, we use \textit{``dpr-ctx\_encoder-single-nq-base''} and \textit{``dpr-question\_encoder-single-nq-base''} for context and tag encoders respectively.


\section{Experiments}

\subsection{Experimental Setup}
\label{sec:experiments-datasets}
We conduct experiments on the following datasets: ARC~\cite{xu2019multi}, QC-Science~\cite{mohania2021tagrec}, and EURLEX57K~\cite{chalkidis2019large}. Details of datasets, metrics, and training details are in Appendix.

For comparison, in addition to simple baselines, we employ some state-of-the-art methods including BERT (prototype)~\cite{snell2017prototypical}, TagRec~\cite{mohania2021tagrec}, TagRec++~\cite{viswanathan2022tagrec++}, and Poly-encoder \cite{humeau2019poly}.
For ablations, built on the bi-encoder (BERT) method, we present three variants: Bi-encoder (BERT) + \ours, Bi-encoder (DPR), and Bi-encoder (DPR) + \ours, where the comparisons between the variants could highlight the contribution of transfer learning and \ours.


\subsection{Results and Analysis}

\textbf{Overall Accuracy:}
The main objective of this work is to improve the bi-encoder models for the purpose of better text classification in two aspects: transfer learning and \ours.
Regarding the effect of using two different pretrained models, 
the results from Table~\ref{tab:main_result} show that models trained on DPR achieve higher performance than models from BERT. 
Specifically, Bi-encoder (DPR) outperforms the Bi-encoder (BERT) for ARC (0.54 > 0.51 in R@1) and QC-Science (0.69 > 0.67 in R@1). 
The performance of the EURLEX57K datasets in both RP@5 and nDCG@5 increases by 0.02. 
Applying our augmentation method to the Bi-encoder (both vanilla BERT and QA-finetuned BERT) improves the performance by 0.06, 0.02, and 0.03 points in ARC, QC-Science, and EURLEX57k, respectively. 
Additionally, the Bi-encoder (DPR) + \ours~demonstrates the highest overall performance in most cases (except for R@3 and R@5 of the QC-Science dataset where differences were small). 
For example, compared to TagRec++, which is the current state-of-the-art model on the datasets, we observed that our best model improves on TagRec++ by 0.05 points in R@1\footnote{We discuss Poly-encoder's low performance in Appendix \ref{sec:appendix_extra_result}.}. 
Figure \ref{fig:EURLEX_dprK} further demonstrates the change in RP@K and nDCG@K across a varying range of values for $K$ on EURLEX57K, where \ours~shows consistently better performance.
Notably, the gap from Bi-encoder (BERT) increases as K increases for both metrics. 

\begin{figure}[!t]
\centering
\begin{subfigure}{\textwidth}
\centering
\includegraphics[width=0.8\linewidth]{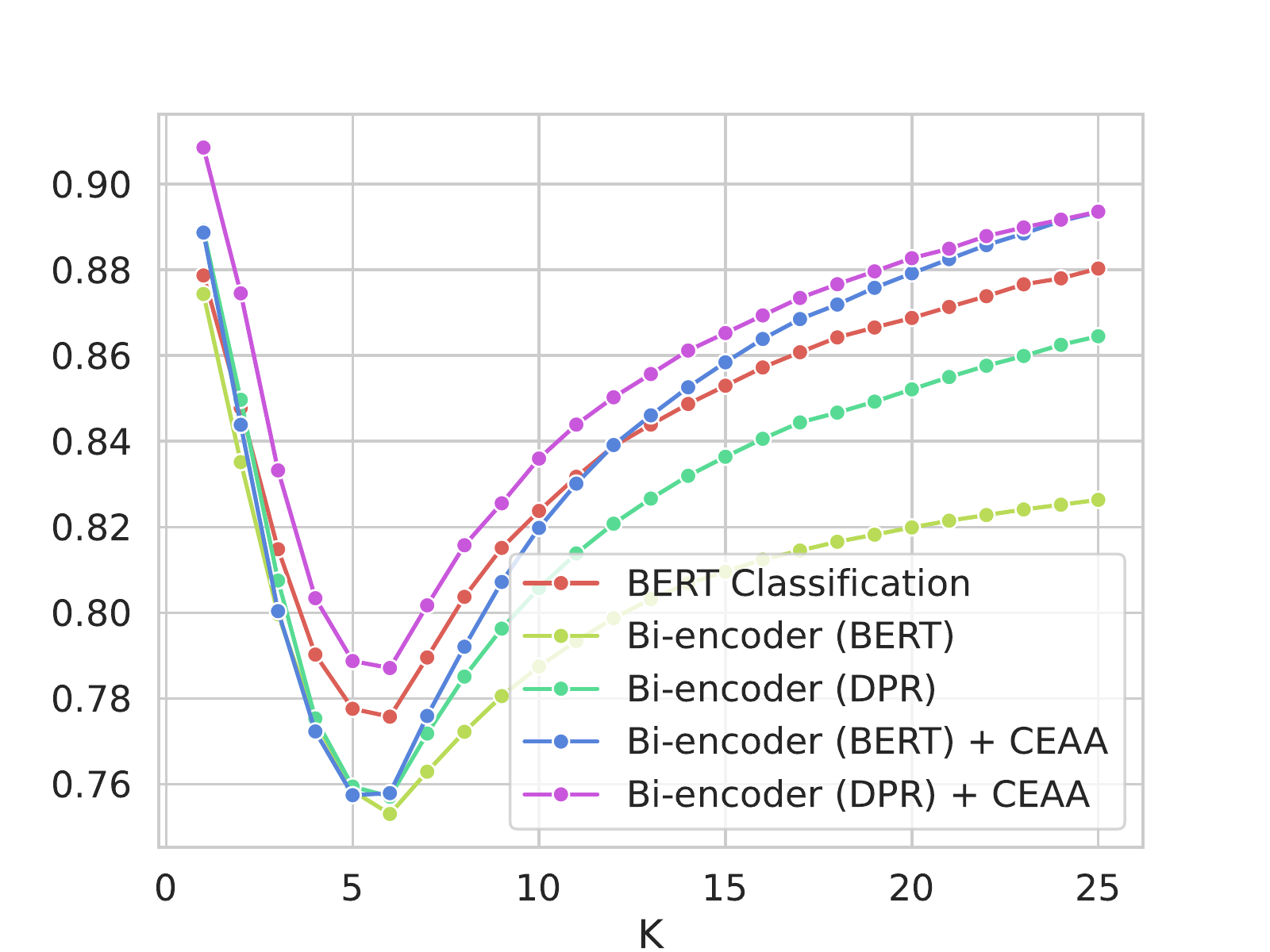}
\caption{RP@K}
\end{subfigure}
\begin{subfigure}{\textwidth}
\centering
\includegraphics[width=0.8\linewidth]{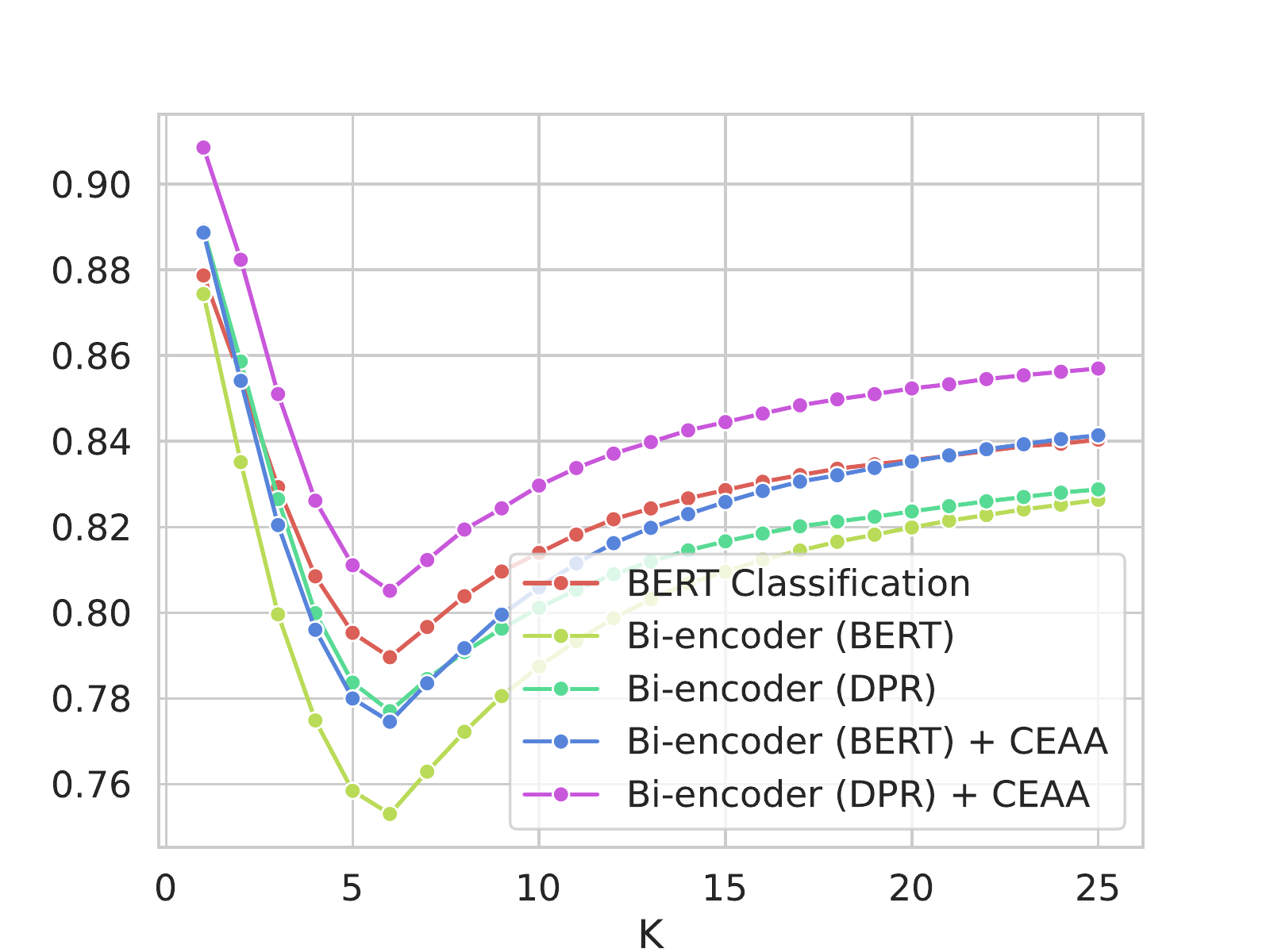}
\caption{nDCG@K}
\end{subfigure}
\caption{Comparison of models on EURLEX57K with two different metrics.}
\label{fig:EURLEX_dprK}
\end{figure}

\begin{figure}[!t]
    \centering
    \includegraphics[width=0.8\columnwidth]{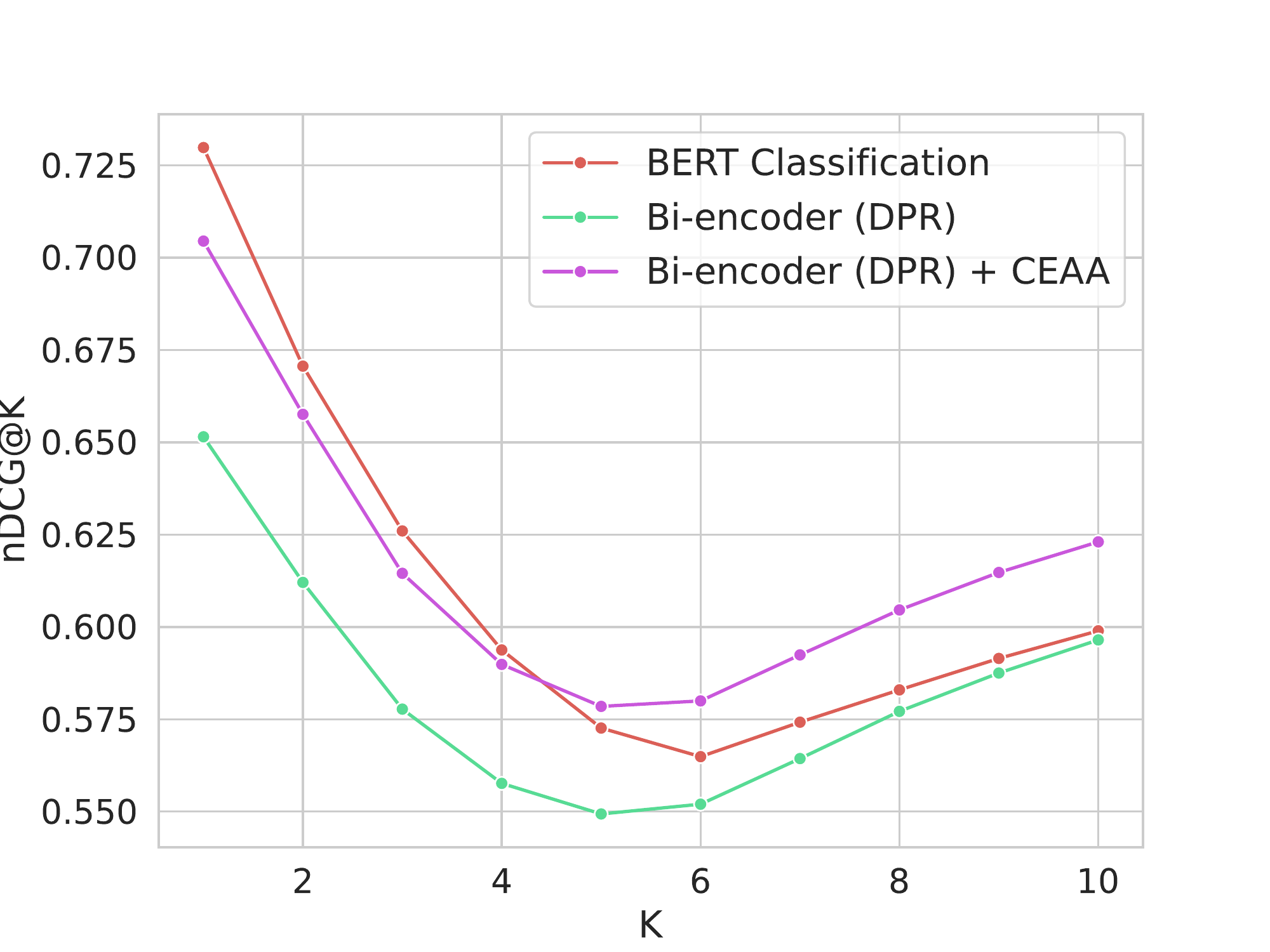}
    \caption{Multi-label evaluation. All models are trained on the single-label version of EURLEX57K but evaluated as multi-label.}
    \label{fig:multi_label}
\end{figure}

\textbf{Multi-label Generalization:}
To further highlight differences between single-label and multi-label settings, the two best models, Bi-encoder (DPR) and Bi-encoder (DPR) + \ours, were trained with a modified single-labeled EURLEX57K dataset, where we sampled only a single tag from the multi-label space. 
When the models are evaluated on the original multi-label dataset, as a context in the EURLEX57K dataset has $\geq 5$ gold tags on average, it is important to achieve high nDCG@K performance on $K \geq 5$. 
The results are presented in Figure~\ref{fig:multi_label}. 
We observe that the models show comparable performance with values of 0.65, 0.70, and 0.73 for Bi-encoder (DPR), Bi-encoder (DPR) + \ours~and BERT classification, respectively at $K=1$. 
Though the classification model performs slightly better than \ours~at low $K$ values, performance significantly degrades for $K \geq 5$. 
Overall, the cross-encoder augmentation helped the model to better find related tags at the top rank.
From these results, we argue that evaluating against the single-labeled dataset may not be an appropriate testing tool for comparing the auto-tagging models, as BERT classification was considered the best at first, even though it is poorly working on multi-label scenarios. 
This problem is critical as multi-label issues are prevalent in education. 

Specifically, we manually checked failure cases of both Bi-encoder (DPR) and Bi-encoder (DPR) + \ours~at top 1, to qualitatively examine which one is better at ranking the relevant tags.
The results in Appendix~\ref{sec:appendix_qualitative} show that Bi-encoder (DPR) + \ours~is able to retrieve better candidates than the Bi-encoder (DPR) more often. 
An interesting example is, given the context \texttt{[``The sector in which employees have more job security is an organized sector'']}, where the gold tag is one related to the economy, the Bi-encoder (DPR) + \ours~returns a tag \texttt{[``human resources'']}, which is sufficiently relevant but not labeled one. 
From these results, we once again confirm that the multi-label problem is severe in the auto-tagging tasks and that our model yields sufficiently significant results beyond the reported performance. 

\begin{figure}[!t]
\centering

\begin{subfigure}{0.8\textwidth}
\centering
\includegraphics[width=\linewidth]{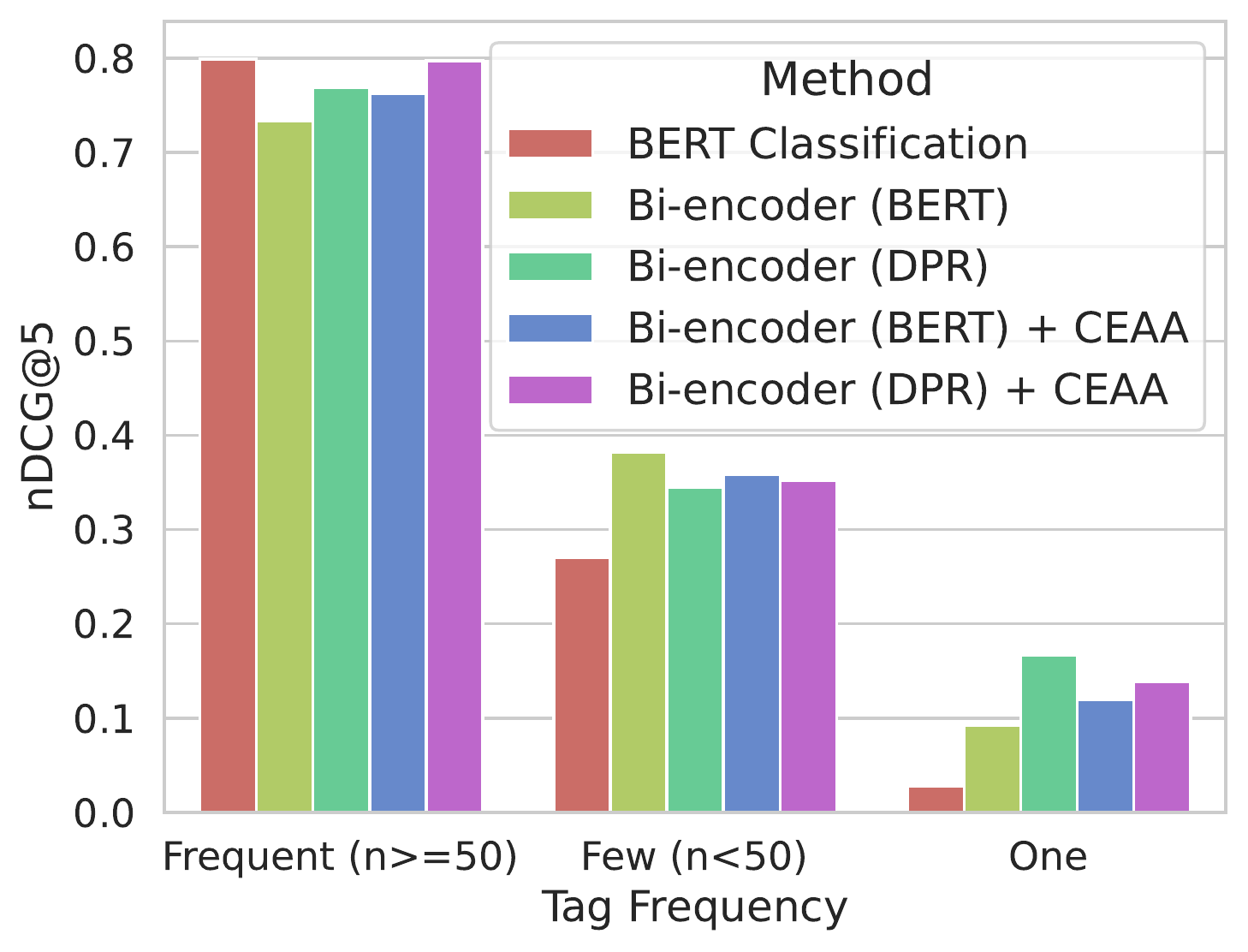}
\vspace{-4mm}
\subcaption{EURLEX57K}
\label{fig:nDCG_EURLEX_Few}
\end{subfigure}

\begin{subfigure}{0.8\textwidth}
\centering
\includegraphics[width=\linewidth]{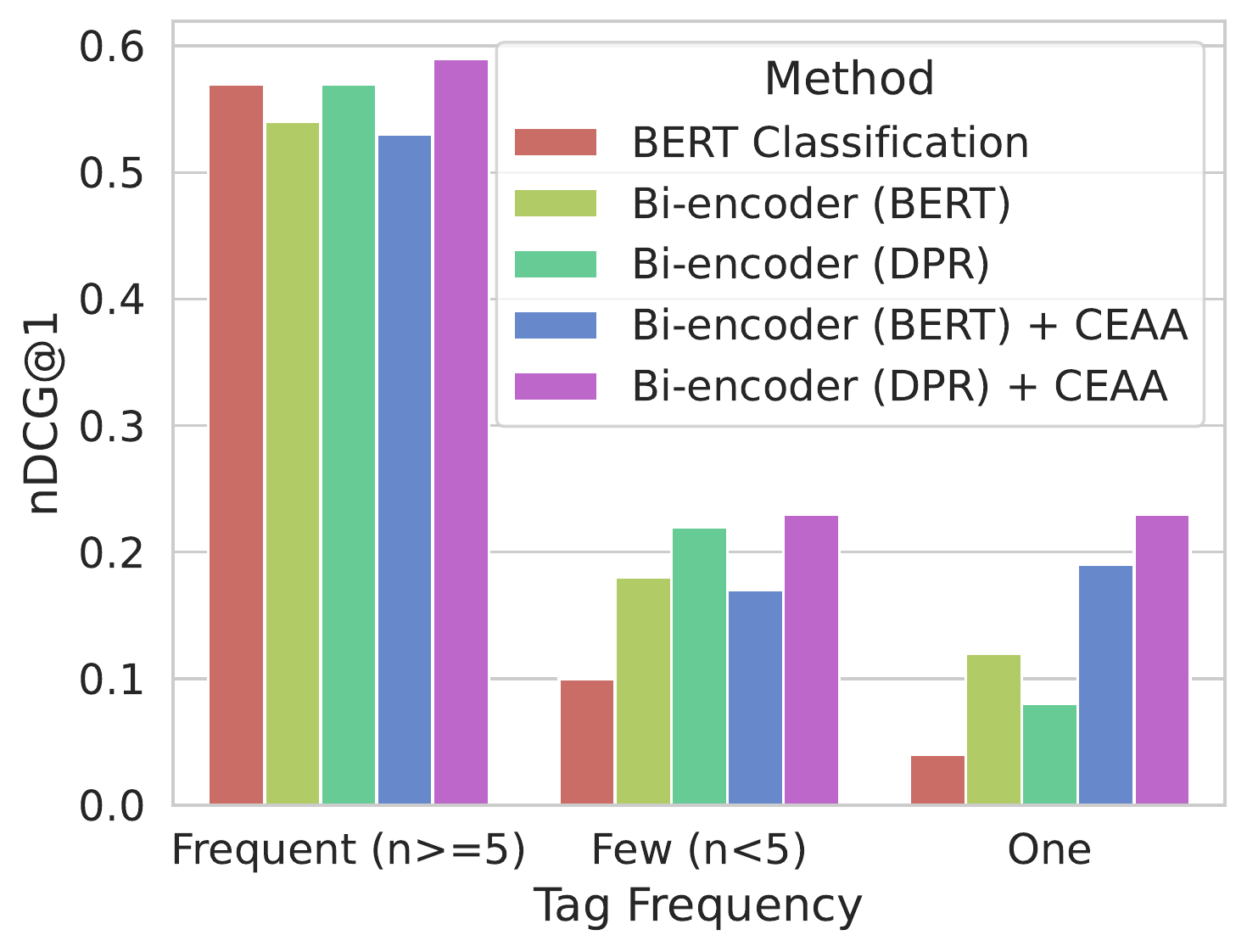}
\vspace{-4mm}
\subcaption{ARC}
\label{fig:nDCG_ARC_Few}
\end{subfigure}

\caption{Analysis on data efficiency. We report nDCG on a varying number of training labels on EURLEX57K and ARC.}
\label{fig:FEW}
\end{figure}

\textbf{Data Efficiency}: To identify the effectiveness of augmentation with low-resource labels, we measured nDCG@5 on the splits of labels based on their occurrence in training data. EURLEX57 considered the labels that occurred more than 50 times in the training set as frequent and few otherwise. We set the ARC dataset's threshold to 5. 
Figure \ref{fig:FEW} shows that both CEAA and transfer learning contribute to better performance for the frequent labels. Further, we observe that the retrieval methods are more effective for the rarely occurring tags than standard classification methods. Notably, in ARC of a smaller dataset than EURLEX57K (5K < 45K), the combination of CEAA and transfer learning, \ours~(DPR), achieves the best performance.

\label{sec:experiments}
\section{Conclusion}
\label{sec:conclusion}

In this paper, we discuss the problem of `\textit{auto-tagging}' with regard to data scarcity due to its large label space - an issue that is critical in the education domain, but also for other domains with a multi-label structure such as jurisdictional or clinical contexts. 
We propose two innovations to address this problem: 
First, exploiting the knowledge of language models trained on large question-answering datasets. 
Second, applying a novel augmentation for bi-encoder architecture inspired by cross-encoders to better capture the full interaction between inputs and labels while maintaining the bi-encoder's efficiency. 
A set of experiments demonstrated the effectiveness of our approach, especially in the multi-label setting.
Future research will explore re-ranking scenarios in which the bi-encoder trained with our cross-encoding augmentation (CEAA) is re-used to effectively re-rank the tags with cross-encoding mechanism as in~\cite{nogueira2019passage}. 


\section{Limitations}
\label{sec:limitations}

\subsection{Limited Size of Language Models}
Due to the recent successes of generative large language models as zero-shot (or, few-shot) text classifiers~\cite{radford2019language,brown2020language}, one may ask about the practicality of our methods. 
Even when disregarding computational efficiency\footnote{Nevertheless, we believe that actionable language models should keep efficiency as one of their core criteria. }, we argue that applying such large language models for XMC problems is not trivial, as it is challenging to constrain the label space appropriately. 
For example, even when the tag candidates we wanted for a task were \texttt{entailment}, \texttt{neutral}, and \texttt{contradiction}), the generative model will output tags outside this range such as \texttt{hamburger}~\cite{raffel2020exploring}. 
In-context learning~\cite{min2022rethinking} may alleviate this concern, but in the context of the large label spaces of our application, the token limits of standard language models will be exceeded.

\subsection{Lack of Knowledge-level Auto-tagging}
Though we pursue text classification tasks in the education domain, the classes usually represent only superficial information, such as chapter titles, which neglects the deeper relationships between educational contents like precondition between knowledge. 
For example, to solve a quadratic problem mathematical problem, an ability to solve the first-order problem is required. 
However, the available texts have only the last superficial tags. 
These concerns were not considered when creating these public datasets. 
Instructor-driven labeling would be an effective and practical solution for knowledge-level auto-tagging.

\subsection{Inefficiency of Tag Encoder}
One may argue that the performance of one BERT system is good enough to cast doubt on using two BERTs for the bi-encoder. In this context, experiments showed additional efficiency of our approach for low-frequency tags. Nonetheless, the current tag encoder could be made much more efficient using a smaller number of layers in BERT which will be explored in the future.

\section{Ethical Considerations}

    Incorrect or hidden decision processes of the AI tagging model could result in the wrong learning path. The system would therefore need to be subject to human monitoring for occasional supervision.
    At the same time, the potential benefits of properly-tagged content will be large for both the learner's learning experience and the teacher's labeling cost as the model can narrow down full tag space to the top-K candidates.

\label{sec:ethical}

    
\label{sec:Acknowledgments}
\bibliography{acl2023}

\begin{thebibliography}{36}
\expandafter\ifx\csname natexlab\endcsname\relax\def\natexlab#1{#1}\fi

\bibitem[{Brown et~al.(2020)Brown, Mann, Ryder, Subbiah, Kaplan, Dhariwal,
  Neelakantan, Shyam, Sastry, Askell et~al.}]{brown2020language}
Tom Brown, Benjamin Mann, Nick Ryder, Melanie Subbiah, Jared~D Kaplan, Prafulla
  Dhariwal, Arvind Neelakantan, Pranav Shyam, Girish Sastry, Amanda Askell,
  et~al. 2020.
\newblock Language models are few-shot learners.
\newblock \emph{Advances in neural information processing systems},
  33:1877--1901.

\bibitem[{Chalkidis et~al.(2020)Chalkidis, Fergadiotis, Kotitsas, Malakasiotis,
  Aletras, and Androutsopoulos}]{chalkidis2020empirical}
Ilias Chalkidis, Manos Fergadiotis, Sotiris Kotitsas, Prodromos Malakasiotis,
  Nikolaos Aletras, and Ion Androutsopoulos. 2020.
\newblock An empirical study on large-scale multi-label text classification
  including few and zero-shot labels.
\newblock \emph{arXiv preprint arXiv:2010.01653}.

\bibitem[{Chalkidis et~al.(2019)Chalkidis, Fergadiotis, Malakasiotis, and
  Androutsopoulos}]{chalkidis2019large}
Ilias Chalkidis, Manos Fergadiotis, Prodromos Malakasiotis, and Ion
  Androutsopoulos. 2019.
\newblock Large-scale multi-label text classification on eu legislation.
\newblock \emph{arXiv preprint arXiv:1906.02192}.

\bibitem[{Chang et~al.(2019)Chang, Yu, Zhong, Yang, and Dhillon}]{chang2019x}
Wei-Cheng Chang, Hsiang-Fu Yu, Kai Zhong, Yiming Yang, and Inderjit Dhillon.
  2019.
\newblock X-bert: extreme multi-label text classification with bert.
\newblock \emph{arXiv preprint arXiv:1905.02331}.

\bibitem[{Choi et~al.(2022)Choi, Jeong, Han, and Hwang}]{choi2022c2l}
Seungtaek Choi, Myeongho Jeong, Hojae Han, and Seung-won Hwang. 2022.
\newblock C2l: Causally contrastive learning for robust text classification.
\newblock In \emph{Proceedings of the AAAI Conference on Artificial
  Intelligence}, volume~36, pages 10526--10534.

\bibitem[{Dekel and Shamir(2010)}]{dekel2010multiclass}
Ofer Dekel and Ohad Shamir. 2010.
\newblock Multiclass-multilabel classification with more classes than examples.
\newblock In \emph{Proceedings of the Thirteenth International Conference on
  Artificial Intelligence and Statistics}, pages 137--144. JMLR Workshop and
  Conference Proceedings.

\bibitem[{Devlin et~al.(2018)Devlin, Chang, Lee, and
  Toutanova}]{devlin2018bert}
Jacob Devlin, Ming-Wei Chang, Kenton Lee, and Kristina Toutanova. 2018.
\newblock Bert: Pre-training of deep bidirectional transformers for language
  understanding.
\newblock \emph{arXiv preprint arXiv:1810.04805}.

\bibitem[{Goel et~al.(2022)Goel, Sahnan, Venktesh, Sharma, Dwivedi, and
  Mohania}]{goel2022k}
Vasu Goel, Dhruv Sahnan, V~Venktesh, Gaurav Sharma, Deep Dwivedi, and Mukesh
  Mohania. 2022.
\newblock K-12bert: Bert for k-12 education.
\newblock In \emph{International Conference on Artificial Intelligence in
  Education}, pages 595--598. Springer.

\bibitem[{Humeau et~al.(2019)Humeau, Shuster, Lachaux, and
  Weston}]{humeau2019poly}
Samuel Humeau, Kurt Shuster, Marie-Anne Lachaux, and Jason Weston. 2019.
\newblock Poly-encoders: Transformer architectures and pre-training strategies
  for fast and accurate multi-sentence scoring.
\newblock \emph{arXiv preprint arXiv:1905.01969}.

\bibitem[{Karpukhin et~al.(2020)Karpukhin, O{\u{g}}uz, Min, Lewis, Wu, Edunov,
  Chen, and Yih}]{karpukhin2020dense}
Vladimir Karpukhin, Barlas O{\u{g}}uz, Sewon Min, Patrick Lewis, Ledell Wu,
  Sergey Edunov, Danqi Chen, and Wen-tau Yih. 2020.
\newblock Dense passage retrieval for open-domain question answering.
\newblock \emph{arXiv preprint arXiv:2004.04906}.

\bibitem[{Koedinger et~al.(2012)Koedinger, Corbett, and
  Perfetti}]{koedinger2012knowledge}
Kenneth~R Koedinger, Albert~T Corbett, and Charles Perfetti. 2012.
\newblock The knowledge-learning-instruction framework: Bridging the
  science-practice chasm to enhance robust student learning.
\newblock \emph{Cognitive science}, 36(5):757--798.

\bibitem[{Kwiatkowski et~al.(2019)Kwiatkowski, Palomaki, Redfield, Collins,
  Parikh, Alberti, Epstein, Polosukhin, Devlin, Lee
  et~al.}]{kwiatkowski2019natural}
Tom Kwiatkowski, Jennimaria Palomaki, Olivia Redfield, Michael Collins, Ankur
  Parikh, Chris Alberti, Danielle Epstein, Illia Polosukhin, Jacob Devlin,
  Kenton Lee, et~al. 2019.
\newblock Natural questions: A benchmark for question answering research.
\newblock \emph{Transactions of the Association for Computational Linguistics},
  7:452--466.

\bibitem[{Lee et~al.(2019)Lee, Chang, and Toutanova}]{lee2019latent}
Kenton Lee, Ming-Wei Chang, and Kristina Toutanova. 2019.
\newblock Latent retrieval for weakly supervised open domain question
  answering.
\newblock \emph{arXiv preprint arXiv:1906.00300}.

\bibitem[{Lin et~al.(2021)Lin, Ma, Lin, Yang, Pradeep, and
  Nogueira}]{lin2021pyserini}
Jimmy Lin, Xueguang Ma, Sheng-Chieh Lin, Jheng-Hong Yang, Ronak Pradeep, and
  Rodrigo Nogueira. 2021.
\newblock Pyserini: An easy-to-use python toolkit to support replicable ir
  research with sparse and dense representations.
\newblock \emph{arXiv preprint arXiv:2102.10073}.

\bibitem[{Lu et~al.(2020)Lu, Du, Liu, and Dipnall}]{lu2020multi}
Jueqing Lu, Lan Du, Ming Liu, and Joanna Dipnall. 2020.
\newblock Multi-label few/zero-shot learning with knowledge aggregated from
  multiple label graphs.
\newblock \emph{arXiv preprint arXiv:2010.07459}.

\bibitem[{Manning(2008)}]{manning2008introduction}
Christopher~D Manning. 2008.
\newblock \emph{Introduction to information retrieval}.
\newblock Syngress Publishing,.

\bibitem[{Medini et~al.(2019)Medini, Huang, Wang, Mohan, and
  Shrivastava}]{medini2019extreme}
Tharun Kumar~Reddy Medini, Qixuan Huang, Yiqiu Wang, Vijai Mohan, and Anshumali
  Shrivastava. 2019.
\newblock Extreme classification in log memory using count-min sketch: A case
  study of amazon search with 50m products.
\newblock \emph{Advances in Neural Information Processing Systems}, 32.

\bibitem[{Min et~al.(2022)Min, Lyu, Holtzman, Artetxe, Lewis, Hajishirzi, and
  Zettlemoyer}]{min2022rethinking}
Sewon Min, Xinxi Lyu, Ari Holtzman, Mikel Artetxe, Mike Lewis, Hannaneh
  Hajishirzi, and Luke Zettlemoyer. 2022.
\newblock Rethinking the role of demonstrations: What makes in-context learning
  work?
\newblock \emph{arXiv preprint arXiv:2202.12837}.

\bibitem[{Mohania et~al.(2021)Mohania, Goyal et~al.}]{mohania2021tagrec}
Mukesh Mohania, Vikram Goyal, et~al. 2021.
\newblock Tagrec: Automated tagging of questions with hierarchical learning
  taxonomy.
\newblock \emph{arXiv preprint arXiv:2107.10649}.

\bibitem[{Nogueira and Cho(2019)}]{nogueira2019passage}
Rodrigo Nogueira and Kyunghyun Cho. 2019.
\newblock Passage re-ranking with bert.
\newblock \emph{arXiv preprint arXiv:1901.04085}.

\bibitem[{Nogueira et~al.(2019{\natexlab{a}})Nogueira, Yang, Cho, and
  Lin}]{nogueira2019multi}
Rodrigo Nogueira, Wei Yang, Kyunghyun Cho, and Jimmy Lin. 2019{\natexlab{a}}.
\newblock Multi-stage document ranking with bert.
\newblock \emph{arXiv preprint arXiv:1910.14424}.

\bibitem[{Nogueira et~al.(2019{\natexlab{b}})Nogueira, Yang, Lin, and
  Cho}]{nogueira2019document}
Rodrigo Nogueira, Wei Yang, Jimmy Lin, and Kyunghyun Cho. 2019{\natexlab{b}}.
\newblock Document expansion by query prediction.
\newblock \emph{arXiv preprint arXiv:1904.08375}.

\bibitem[{Qu et~al.(2021)Qu, Ding, Liu, Liu, Ren, Zhao, Dong, Wu, and
  Wang}]{qu2021rocketqa}
Yingqi Qu, Yuchen Ding, Jing Liu, Kai Liu, Ruiyang Ren, Wayne~Xin Zhao, Daxiang
  Dong, Hua Wu, and Haifeng Wang. 2021.
\newblock Rocketqa: An optimized training approach to dense passage retrieval
  for open-domain question answering.
\newblock In \emph{Proceedings of the 2021 Conference of the North American
  Chapter of the Association for Computational Linguistics: Human Language
  Technologies}, pages 5835--5847.

\bibitem[{Radford et~al.(2019)Radford, Wu, Child, Luan, Amodei, Sutskever
  et~al.}]{radford2019language}
Alec Radford, Jeffrey Wu, Rewon Child, David Luan, Dario Amodei, Ilya
  Sutskever, et~al. 2019.
\newblock Language models are unsupervised multitask learners.

\bibitem[{Raffel et~al.(2020)Raffel, Shazeer, Roberts, Lee, Narang, Matena,
  Zhou, Li, Liu et~al.}]{raffel2020exploring}
Colin Raffel, Noam Shazeer, Adam Roberts, Katherine Lee, Sharan Narang, Michael
  Matena, Yanqi Zhou, Wei Li, Peter~J Liu, et~al. 2020.
\newblock Exploring the limits of transfer learning with a unified text-to-text
  transformer.
\newblock \emph{J. Mach. Learn. Res.}, 21(140):1--67.

\bibitem[{Ren et~al.(2021)Ren, Qu, Liu, Zhao, She, Wu, Wang, and
  Wen}]{ren2021rocketqav2}
Ruiyang Ren, Yingqi Qu, Jing Liu, Wayne~Xin Zhao, Qiaoqiao She, Hua Wu, Haifeng
  Wang, and Ji-Rong Wen. 2021.
\newblock Rocketqav2: A joint training method for dense passage retrieval and
  passage re-ranking.
\newblock \emph{arXiv preprint arXiv:2110.07367}.

\bibitem[{Snell et~al.(2017)Snell, Swersky, and Zemel}]{snell2017prototypical}
Jake Snell, Kevin Swersky, and Richard Zemel. 2017.
\newblock Prototypical networks for few-shot learning.
\newblock \emph{Advances in neural information processing systems}, 30.

\bibitem[{Thakur et~al.(2020)Thakur, Reimers, Daxenberger, and
  Gurevych}]{thakur2020augmented}
Nandan Thakur, Nils Reimers, Johannes Daxenberger, and Iryna Gurevych. 2020.
\newblock Augmented sbert: Data augmentation method for improving bi-encoders
  for pairwise sentence scoring tasks.
\newblock \emph{arXiv preprint arXiv:2010.08240}.

\bibitem[{Urbanek et~al.(2019)Urbanek, Fan, Karamcheti, Jain, Humeau, Dinan,
  Rockt{\"a}schel, Kiela, Szlam, and Weston}]{urbanek2019learning}
Jack Urbanek, Angela Fan, Siddharth Karamcheti, Saachi Jain, Samuel Humeau,
  Emily Dinan, Tim Rockt{\"a}schel, Douwe Kiela, Arthur Szlam, and Jason
  Weston. 2019.
\newblock Learning to speak and act in a fantasy text adventure game.
\newblock \emph{arXiv preprint arXiv:1903.03094}.

\bibitem[{Vig and Ramea(2019)}]{vig2019comparison}
Jesse Vig and Kalai Ramea. 2019.
\newblock Comparison of transfer-learning approaches for response selection in
  multi-turn conversations.
\newblock In \emph{Workshop on DSTC7}.

\bibitem[{Viswanathan et~al.(2022)Viswanathan, Mohania, and
  Goyal}]{viswanathan2022tagrec++}
Venktesh Viswanathan, Mukesh Mohania, and Vikram Goyal. 2022.
\newblock Tagrec++: Hierarchical label aware attention network for question
  categorization.
\newblock \emph{arXiv preprint arXiv:2208.05152}.

\bibitem[{Wolf et~al.(2020)Wolf, Debut, Sanh, Chaumond, Delangue, Moi, Cistac,
  Rault, Louf, Funtowicz et~al.}]{wolf2020transformers}
Thomas Wolf, Lysandre Debut, Victor Sanh, Julien Chaumond, Clement Delangue,
  Anthony Moi, Pierric Cistac, Tim Rault, R{\'e}mi Louf, Morgan Funtowicz,
  et~al. 2020.
\newblock Transformers: State-of-the-art natural language processing.
\newblock In \emph{Proceedings of the 2020 conference on empirical methods in
  natural language processing: system demonstrations}, pages 38--45.

\bibitem[{Wolf et~al.(2019)Wolf, Sanh, Chaumond, and
  Delangue}]{wolf2019transfertransfo}
Thomas Wolf, Victor Sanh, Julien Chaumond, and Clement Delangue. 2019.
\newblock Transfertransfo: A transfer learning approach for neural network
  based conversational agents.
\newblock \emph{arXiv preprint arXiv:1901.08149}.

\bibitem[{Xu et~al.(2019)Xu, Jansen, Martin, Xie, Yadav, Madabushi, Tafjord,
  and Clark}]{xu2019multi}
Dongfang Xu, Peter Jansen, Jaycie Martin, Zhengnan Xie, Vikas Yadav,
  Harish~Tayyar Madabushi, Oyvind Tafjord, and Peter Clark. 2019.
\newblock Multi-class hierarchical question classification for multiple choice
  science exams.
\newblock \emph{arXiv preprint arXiv:1908.05441}.

\bibitem[{Zhang et~al.(2022)Zhang, Wang, Yang, Yu, Vu, and Lei}]{zhang2022long}
Ruohong Zhang, Yau-Shian Wang, Yiming Yang, Donghan Yu, Tom Vu, and Likun Lei.
  2022.
\newblock Long-tailed extreme multi-label text classification with generated
  pseudo label descriptions.
\newblock \emph{arXiv preprint arXiv:2204.00958}.

\bibitem[{Zhang et~al.(2015)Zhang, Zhao, and LeCun}]{zhang2015character}
Xiang Zhang, Junbo Zhao, and Yann LeCun. 2015.
\newblock Character-level convolutional networks for text classification.
\newblock \emph{Advances in neural information processing systems}, 28.

\end{thebibliography}
\bibliographystyle{acl_natbib}


\appendix
\label{sec:appendix}

\section{Experimental Setup}
\subsection{Data Statistics}
\label{sec:appendix_data}

    \textbf{ARC}~\cite{xu2019multi}: This dataset consists of 7,775  multiple-choice questions and answer pairs from the science domain. Each data is paired with classification taxonomy. The taxonomy is constructed to categorize questions into coarse to fine chapters in a science exam.
    There are a total of 420 unique labels. The dataset is split in train, validation, and test by 5,597, 778, and 1400 samples.
       
   \textbf{QC-Science}~\cite{mohania2021tagrec}: this larger dataset consists of 47,832 question-answer pairs also in the science domain  with 312 unique tags. Each tags are hierarchical labels in the form of subject, chapter, and topic. The train, validation, and test sets consist of 40,895, 2,153, and 4,784 samples.
   
  \textbf{EURLEX57K}~\cite{chalkidis2019large}: The dataset contains 57,000 English legislative documents from EUR-LEX with a split of 45,000, 6,000, and 6,000. Every document is tagged with multi-label concepts from European Vocabulary. The average number of tags per document is 5, totaling 4,271 tags. Additionally, the  dataset divides the tags into frequent (746), few (3,362), and zero (163), based on whether they appeared more than 50, fewer than 50, but at least once, or never, respectively. 
  
 
    
\subsection{Details on Evaluation Metric}
\label{sec:appendix_metric}
    In this section, we explain the metric used in the paper. First, recall@K($R@K$) is calculated as follows:
    
    \begin{equation}
    R@K =\frac{1}{N}\sum_{n=1}^{N}\frac{S_t(K)}{R_n}
    \end{equation}
    
    where $N$ is the number of samples to test, $R_n$ is the number of true tags for a sample $n$, and $S_t(K)$ is the number of true tags within the top-$K$ results. 
    For evaluation on multi-label dataset we used R-Precision@K ($RP@K$) \cite{chalkidis2019large}: 
    \begin{equation}
    RP@K =\frac{1}{N}\sum_{n=1}^{N}\frac{S_t(K)}{min(R_n,K)}
    \end{equation}
    RP@K divides the number of true positives within $K$ by the minimum value between $K$ and $R_n$, resulting in a more fair and informative comparison in a multi-label setting. 

    nDCG@K \cite{manning2008introduction} is another metric commonly used in such tasks. The difference between RP@K and nDCG@K is the latter includes the ranking quality by accounting for the location of the relevant tags within the top-K retrieved tags as follows: 
    
    \begin{equation}
    nDCG@K =\frac{1}{N}\sum_{n=1}^{N}Z_{K_n}\sum_{k=1}^{K}\frac{Rel(n,k)}{log_2(1+k)}
    \end{equation}
            
    where $Rel(n,k)$ is the relevance score given by the dataset between a retrieved tag $k$ of a sample $n$. The value can be different if the tags' relevant score is uniquely given by the dataset. Without extra information, it is always one if relevant and zero otherwise. $Z_{K_n}$ is a normalizing constant that is output of DCG@K when the optimal top-K were retrieved as true tags.
    
\subsection{Hyperparmeter Setting} 
\label{sec:appendix_training}
The architecture we used can handle a maximum of 512 tokens. Therefore, to concatenate tag tokens with context tokens, we set the maximum context token to 490 and truncate if the context is longer. The remaining space is used for tag token concatenation.  For every dataset, we used 20 contexts inside a batch. The number of unique tags inside a batch can vary with multi-label settings. During cross-encoder augmentation, we sampled five negative tags for each context to be joined together and one positive tag. We used Adam optimizer with a learning rate of 1e-5. For inference, we used the Pyserini framework to index the entire tag set embeddings \cite{lin2021pyserini}.

\section{Additional Results and Comments}


\subsection{Comments on Poly-encoder}
\label{sec:appendix_extra_result}

    In this section, we discuss the low performance of Poly-encoder \cite{humeau2019poly} in our main results. To be more specific, poly-encoder-16 and 360 were found to be performing below TagRec++. The value 16, and 360 is the number of vectors to represent a context. We think the low performance could be due to a potential implementation issue of the poly-encoder into the classification task. The  performance could differ if we had used 16 or 360 vectors to represent the tag rather than a context. For our future work, we also aim to investigate this  change.
\subsection{Extra Qualitative Result}
\label{sec:appendix_qualitative}
    Table \ref{tab:extra qual result} shows the samples we used to find the potential of \ours~method in multi-label tasks. The shown results were randomly picked.
    
\begin{table*}[t]
    \caption{Extra result of sampled QC-Science to show the strength of \ours~method in multi-label tasks. }
    \label{tab:extra qual result}
    \begin{tabular}{l p{0.75\linewidth}}
    \\\toprule
    
    \textbf{Context} & \texttt{A good conductor of heat is a steel ruler.}
    \\
    \textbf{Ground Truth} & \texttt{\textbf{science >> heat}}  
    \\
    \textbf{Bi-Encoder} & \texttt{science >> fun with magnets}
    \\
    \textbf{Bi-Encoder + \ours}  &\texttt{science >> sorting materials into groups}
    \\\hline
    \textbf{Context} & \texttt{The operating system which allows two or more users to run programs at the same time is multi-user.}
    \\
    \textbf{Ground Truth} & \texttt{computer science[c++] >> computer overview}   
    \\
    \textbf{Bi-Encoder} & \texttt{computer science >> introduction to computer}
    \\
    \textbf{Bi-Encoder + \ours} & \texttt{\textbf{computer science[c++]>>working with operating system}}
    \\\hline
    \textbf{Context} & \texttt{The radiation which will deflect in electric field is cathode rays}
    \\
    \textbf{Ground Truth} & \texttt{\textbf{physics >> physics : part - ii >> dual nature of radiation and matter}}   
    \\
    \textbf{Bi-Encoder} & \texttt{physics >> physics : part - ii>>atoms}
    \\
    \textbf{Bi-Encoder + \ours} & \texttt{\textbf{physics >> physics: part - I >> electric charges and fields}}
    \\\hline
    \textbf{Context} & \texttt{What do we call the resources that helps in production process? Factors of Production}
    \\
    \textbf{Ground Truth} & \texttt{social science >> economics >> the story of village palampur}   
    \\
    \textbf{Bi-Encoder} & \texttt{social science >> geography : resource and development>>resources}
    \\
    \textbf{Bi-Encoder + \ours} & \texttt{\textbf{social science >> economics >> people as resource}}
    \\\hline
    \textbf{Context} & \texttt{The Civil Law to protect women against domestic violence was passed in 2006.}
    \\
    \textbf{Ground Truth} & \texttt{\textbf{social science>>civics : social and political life>>judiciary}}   
    \\
    \textbf{Bi-Encoder} & \texttt{social science >> civics : social and political life >> understanding laws}
    \\
    \textbf{Bi-Encoder + \ours} & \texttt{\textbf{social science >> civics : social and political life - ii >> women change the world}}
    \\\hline
    
\textbf{Context} & \texttt{In the mid 18 th century, major portion of eastern India was under the control of the British.}
    \\
    \textbf{Ground Truth} & \texttt{\textbf{social science >> eighteenth-century poltical formations >> the later mughals and the emergence of new states}}   
    \\
    \textbf{Bi-Encoder} & \texttt{\textbf{social science >> history : our pasts - ii >> eighteenth-century political formations}}
    \\
    \textbf{Bi-Encoder + \ours} & \texttt{social science >> history : india and the contemporary world - i >> peasant and farmers}
    \\\hline

    \textbf{Context} & \texttt{Spirogyra is called so because chloroplasts are spiral.}
    \\
    \textbf{Ground Truth} & \texttt{science}   
    \\
    \textbf{Bi-Encoder} & \texttt{\textbf{science>>cell structure and functions}}
    \\
    \textbf{Bi-Encoder + \ours} & \texttt{science >> life processes}
    \\\hline

    \textbf{Context} & \texttt{The element having electronic configuration 2,8,4 is silicon.}
    \\
    \textbf{Ground Truth} & \texttt{\textbf{science >> periodic classification of elements}}   
    \\
    \textbf{Bi-Encoder} & \texttt{chemistry >> chemistry : part I >> the solid state}
    \\
    \textbf{Bi-Encoder + \ours} & \texttt{science >>    structure of the atom}
    \\\hline

    \textbf{Context} & \texttt{The clouds are actually tiny droplets of water.}
    \\
    \textbf{Ground Truth} & \texttt{\textbf{science >> water}}      
    \\
    \textbf{Bi-Encoder} & \texttt{\textbf{science >> air around us}}
    \\
    \textbf{Bi-Encoder + \ours} & \texttt{\textbf{social science >> geography : our environment>>air}}
    \\\bottomrule
    \end{tabular}
\end{table*}



\end{document}